\pdfoutput=1

\documentclass[11pt]{article}
\usepackage{naacl2021}
\usepackage{times}
\usepackage{latexsym}

\usepackage{microtype}
\usepackage{enumerate}
\usepackage[inline]{enumitem}
\usepackage{url}

\usepackage{amsmath}
\usepackage{amssymb}
\usepackage{amsfonts}
\usepackage{booktabs}
\usepackage{supertabular}
\usepackage{multirow}
\usepackage{lipsum}
\usepackage{cleveref}
\usepackage{xfrac}

\crefname{section}{\S}{\S\S}
\Crefname{section}{\S}{\S\S}
\crefname{table}{Tab.}{}
\crefname{figure}{Fig.}{}
\crefname{algorithm}{Algorithm}{}
\crefname{equation}{eq.}{}
\crefname{appendix}{App.}{}
\crefname{thm}{Theorem}{}
\crefname{prop}{Proposition}{}
\crefname{cor}{Corollary}{}
\crefname{observation}{Observation}{}
\crefname{assumption}{Assumption}{}
\crefformat{section}{\S#2#1#3}

\usepackage{csquotes}

\DeclareMathOperator{\Entr}{H}
\DeclareMathOperator{\MI}{I}

\DeclareMathOperator{\PMI}{PMI}
\DeclareMathOperator{\Unc}{U}
\DeclareMathOperator{\softmax}{softmax}
\DeclareMathOperator{\LSTM}{LSTM}

\newcommand{\word}[1]{\textit{#1}}
\newcommand{\phon}[1]{[#1]}
\newcommand{\concept}[1]{\textsc{#1}}
\newcommand{\defn}[1]{\textbf{#1}}
\newcommand{\vw}{\mathbf{w}}
\newcommand{\vv}{\mathbf{v}}
\newcommand{\Entrtheta}{\Entr_{\boldsymbol{\theta}}}
\newcommand{\ptheta}{p_{\boldsymbol{\theta}}}

\newcommand{\wordform}{\mathbf{w}^{(n)}}
\DeclareMathOperator{\wordformspace}{\Sigma^*}
\DeclareMathOperator{\stringending}{\#}

\newcommand\citepossessive[1]{\citeauthor{#1}'s\ (\citeyear{#1})}

\everypar{\looseness=-1}

\title{Finding Concept-specific Biases in Form--Meaning Associations}

\usepackage{tipa}

\newcommand{\ucambridge}{\normalfont \text{\textipa{D}}}
\newcommand{\ethz}{\text{\normalfont \textipa{Q}}}
\newcommand{\google}{\normalfont \text{\textipa{@}}}
\newcommand{\harvard}{\normalfont \text{\textipa{N}}}
\newcommand{\mpi}{\normalfont \text{\textipa{9}}}
\newcommand{\hse}{\normalfont \text{\textipa{6}}}
\newcommand{\kazan}{\normalfont \text{\textipa{R}}}

\author{Tiago Pimentel$^{\ucambridge}$
~ Brian Roark$^{\google}$
~ S{\o}ren Wichmann$^{\kazan}$
~ Ryan Cotterell$^{\ucambridge,\ethz}$ 
~ Dami\'{a}n Blasi$^{\harvard, \mpi, \hse}$ 
\\
\\
  $^{\ucambridge}$University of Cambridge%
  ~\;~\;~\;~\;~\;~ $^{\google}$Google%
  ~\;~\;~\;~\;~\;~ $^{\kazan}$Kazan Federal University%
  ~\;~\;~\;~\;~\;~ $^{\ethz}$ETH Z\"{u}rich~\;~\;~ \\
    $^{\harvard}$Harvard University%
  ~\;~\;~\;~ $^{\mpi}$MPI for Evolutionary Anthropology%
  ~\;~\;~\;~ $^{\hse}$HSE University%
  \\
  \texttt{tp472@cam.ac.uk}%
  ,~\;~ \texttt{roark@google.com}%
  ,~\;~ \texttt{wichmannsoeren@gmail.com} \\
  \texttt{ryan.cotterell@inf.ethz.ch}%
  ,~\;~ \texttt{dblasi@fas.harvard.edu}
}

\date{}

\begin{document}
\maketitle
\begin{abstract}
This work presents an information-theoretic operationalisation
of cross-linguistic non-arbitrariness. It is not a new idea that there are small, cross-linguistic associations between the forms and meanings of words. For instance, it has been claimed \cite{blasi2016sound} that the word for \concept{tongue} is more likely than chance to contain the phone \phon{l}. By controlling for the influence of language family and geographic proximity within a very large concept-aligned, cross-lingual lexicon, we extend methods previously used to detect within language non-arbitrariness \cite{pimentel2019meaning} to measure cross-linguistic associations.
We find that there is a significant effect of non-arbitrariness, but it is unsurprisingly small (less than 0.5\% on average according to our information-theoretic 
estimate). We also provide a concept-level analysis which shows that 
a quarter of the concepts considered in our work exhibit a significant level of cross-linguistic non-arbitrariness. In sum, the paper provides new methods to detect cross-linguistic associations at scale, and confirms their effects are minor.
\end{abstract}

\section{Introduction} \label{sec:introduction}

The arbitrariness of the sign, i.e. the principle that a word's form is unrelated to what it denotes, was one of the cornerstones in the     structuralist revolution in linguistics \cite{saussure1916course}.
While languages do seem to adhere to the principle to a large extent, researchers have repeatedly uncovered evidence that there are preferences in form--meaning matches \citep{perniss2010iconicity}. 
Indeed, the notion that these small, but systematic, form--meaning relations hold \emph{across} the world's languages has become a mainstream topic of research in the last couple of decades.%
\footnote{See \cref{sec:background} below for a brief literature review and \newcite{dingemanse2015arbitrariness} for a more comprehensive one.}

Determining effective metrics to capture meaningful form--meaning associations is far from trivial, though, and researchers have explored a substantial number of statistical and heuristic approaches \cite{bergen2004psychological,wichmann2010sound,johansson2013motivations,haynie2014sound,gutierrez2016finding,blasi2016sound,joo2019phonosemantic}.
Previous studies differ from each other along (at least) three axes:
\begin{enumerate*}[label=(\roman*)]
\item  which unit is used to measure wordform similarity (e.g., phonemes, sub-phonemic features or arbitrary sequences);
\item how they deploy a baseline for statistical comparison (e.g. permute forms with meanings, or propose a generative model that yields wordforms uninformed by their meaning) and 
\item whether they study non-arbitrariness within or across languages.
\end{enumerate*}

\begin{figure}
    \centering
    \includegraphics[width=\columnwidth]{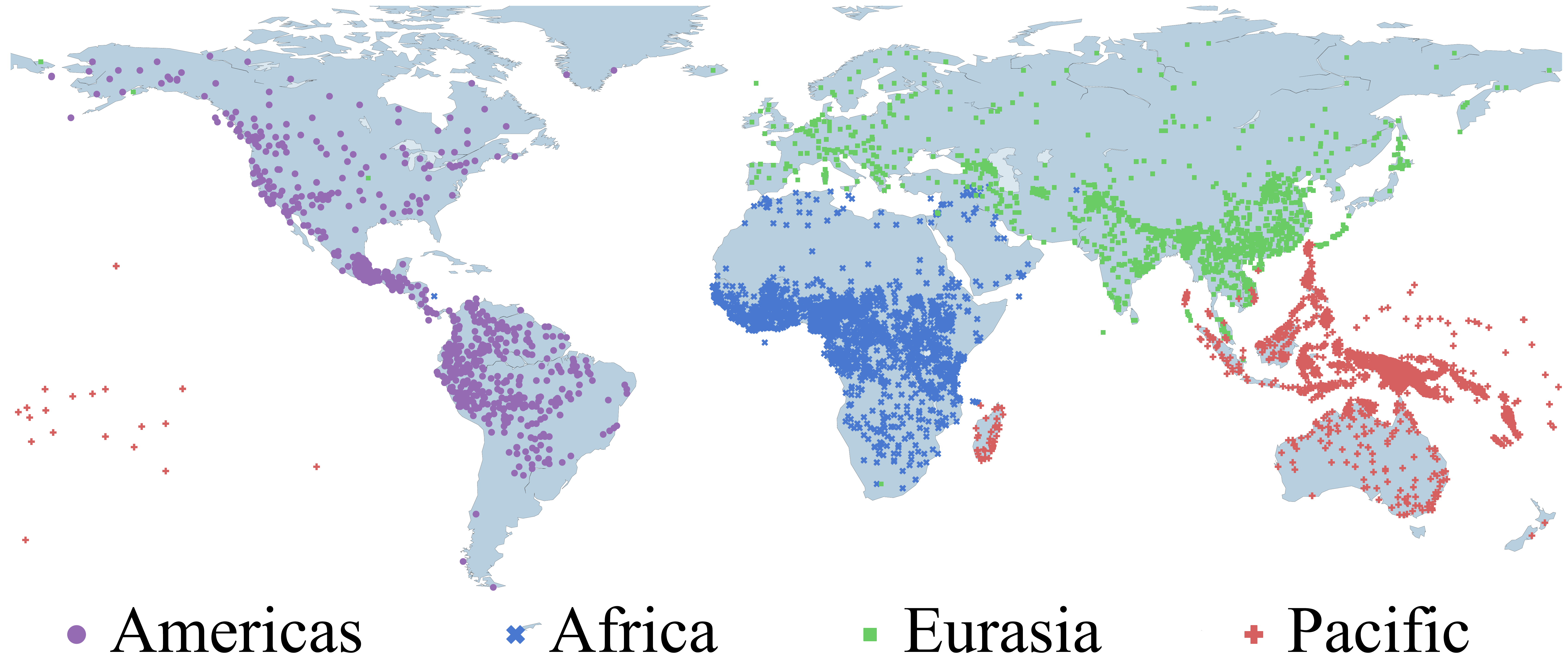}
    \caption{We used a sample of 5189 languages (9148 doculects) to study cross-linguistic systematicity. In this map, the colours represent the four macroareas to which languages were assigned.}
    \label{fig:macroareas}
    \vspace{-5pt}
\end{figure}

\citet{pimentel2019meaning} provide the first holistic measure of non-arbitrariness (in a large vocabulary sample of a single language) using tools from information theory, and apply their measure to discover phonesthemes.\footnote{Phonesthemes are sub-morphemic units which are associated in a language with some small semantic domain.}
Our work extends their approach to the problem of discovering and estimating the strength of frequent \emph{cross-linguistic} form--meaning associations (e.g. iconicity and systematicity) in individual concepts. 
We do this by adapting \citepossessive{pimentel2019meaning} approach, modelling
form--meaning associations in a large collection of basic vocabulary wordlists covering close to $\sfrac{3}{4}$ of the world's languages \citep[see \cref{fig:macroareas} and][]{wichmann2020asjp}.
By taking the words in these lists to be random variables and asking how much information within wordforms is explained by the meaning they refer to, we obtain a quantitative estimate of cross-linguistic non--arbitrariness.\looseness=-1

Specifically, we propose to model a universal (language-independent) form distribution (using neural language models), and then we estimate concept-specific distributions. %
With these in hand, we are able to determine how much the meaning of a concept predicts its form cross-linguistically---by measuring the mutual information between them; see \cref{sec:methods} for details. 
This method further allows us to identify which concepts exhibit stronger non-arbitrary form--meaning association and which form patterns are more likely to occur in them.

In order to maximise the reliability of the observed associations, we implement stringent controls for genealogical and areal effects, as well as for the size of each language family. See \cref{sec:crosslingual-contros} for details on these controls.
After introducing these controls, we find that wordlists display an average of around $0.01$ bits of form--meaning mutual information 
explained by cross-linguistic non-arbitrariness ($\approx 0.3\%$ of the wordform uncertainty) with substantial variation among concepts and languages.\footnote{See \cref{sec:data} for a discussion on potential biases in the dataset that likely influence the actual numerical value derived.} 
Of the 100 basic concepts in our data, we find a statistically identifiable pattern in 26 of them ($p < 0.01$). Inspection of the results show that our method recovers previously proposed associations, e.g. the association of \phon{l} with the concept \concept{tongue} and \phon{p} with \concept{full} \cite{blasi2016sound}. 
\section{Non-Arbitrary Form--Meaning Associations}\label{sec:background}

Several studies have looked at non-arbitrary patterns in languages, be it systematicity \citep{shillcock2001filled,gutierrez2016finding,dautriche2017wordform,pimentel2019meaning} or iconicity \citep{dingemanse2012advances, dingemanse2018redrawing}.
With respect to cross-linguistic non-arbitrariness specifically, 
the hypothesised sources of form--meaning associations range from the fact that humans are endowed with the same neurocognitive architecture \citep{bankieris2015link} to their encountering similar experiences within the world \citep{parise2014natural}.

While global non-arbitrary form--meaning associations have been hypothesised to exist at different levels of linguistic description \citep{haiman1980iconicity}, by far the component of language that has received the most attention in this respect is the lexicon.
A few circumstances facilitate this type of research in contrast to other domains of grammar. For instance, the space of possible words that could be used in a given language to refer to an arbitrary referent is large, whereas the relative canonical order of a verb with respect to its  object complement is substantially smaller (which renders cross-linguistic similarities less informative than in the first case). Additionally, the sheer amount of data available in the form of wordlists exceeds other types of linguistic data for the languages of the world.

As a consequence, some of the largest evaluations of non-arbitrary form--meaning associations involve systematic wordlists with comparable referents across languages \citep{wichmann2010sound,johansson2013motivations,haynie2014sound,blasi2016sound,joo2019phonosemantic}. Most of these studies were focused on the regular association between phonemic or phonetic units with meaning, occasionally controlling for other potential sources of form--meaning association such as phonotactics or word length \citep{blasi2016sound}. 
While useful, the estimates emerging from this type of study can be regarded as lower bounds to the total amount of non-arbitrary associations found in the vocabulary.
Recent efforts have resulted in datasets with thousands of languages \citep{wichmann2020asjp}, with which linguists can look for universal statistical patterns \citep{wichmann2010sound, blasi2016sound}. These studies, though, only looked at the presence (or not) of individual phones in %
words, not accounting for their connections.
Our methods rely on neural phonotactic models, similar to those used by \citet{pimentel2020phonotactic}, thus capturing a broader range of potential correspondences.

\vspace{-4pt}
\section{Data}\label{sec:data}
\vspace{-2pt}

An exceptional resource with substantial cross-linguistic representation is provided in the Automated Similarity Judgment Program, better known by its acronym ASJP \citep{wichmann2020asjp}. ASJP is a collection of basic vocabulary wordlists, i.e. lists of words with referents that are expected to be widely attested across human societies. It involves body parts, some colour terms, lower numerals, general properties (such as big or round), and flora and fauna that are usually found in places where humans live (e.g. trees and dogs). The individual words in ASJP are transcribed by field linguists in a specific phonetic annotation scheme that involves 41 symbols, chosen in order to maximise cross-linguistic utility by merging rare phones with similar phonetic features within the same category. 
These wordlists are assembled with the purpose of studying the history of languages---following the tradition established by \newcite{swadesh1955towards}---under the principles of the comparative method.%

ASJP has gathered, in its latest iterations, data for close to $\sfrac{3}{4}$ of the world's languages, which makes it an unparalleled resource for evaluating form--meaning associations across spoken languages. 
Furthermore, the vocabulary in its wordlists was chosen as so to be resistant to borrowings---making it especially interesting for our purposes of finding universal form--meaning biases. 
We leave out pidgin and creole data,\footnote{Pidgins are believed to rely particularly on iconicity due to their smaller degree of lexicalisation; this reliance then diminishes as it morphs into a creole \citep{romaine1988pidgins}. Future work could expand the methods here to study this phenomenon.}
as defined by the World Atlas of Language Structures \citep{wals}, since they are ambiguous with relation to their genealogical affiliation. 
We also omit constructed and fake languages (e.g. Esperanto and Taensa). This leaves us 9148 doculects (or wordlists) from 5189 languages.\footnote{When there is more than one wordlist for one language (as defined by their ISO-codes) one can sometimes refer to them as different dialects, but these are often just alternative versions of the same language as recorded by different linguists. There can be as much variation in such different recordings as among different dialects recorded by one and the same linguist. For those reasons, it is practical to use the term \emph{doculect}, which we adopt here. This is a neutral term that refers to some dialect \emph{as recorded in some specific source}.}\looseness=-1

Form--meaning associations have been studied in earlier versions  of this dataset. 
Firstly, \citet{wichmann2010sound} studied the average form across different concepts in ASJP, and found a number of tentative patterns pointing to non-arbitrariness. Yet the lack of historical and statistical controls compromised the nature of such patterns: form--meaning associations could be due to widespread linguistic contact (e.g. the word for \concept{dog}, \citealt{pache2016words}) or to its fortuitous presence in large families. \citet{blasi2016sound}, however, provide a conservative evaluation of individual form--meaning associations by imposing a restrictive set of conditions. They looked for associations that were present in a minimum number of continents and language families. This resulted in a sizable number of non-arbitrary associations, many of which had been highlighted as interesting based on behavioural and linguistic experiments in a handful of languages.

\paragraph{Data Disclaimer.} As mentioned above, ASJP gathers lists of wordforms that are expected to be present across most human societies and their corresponding language(s). While this guarantees a fair coverage in our study, it limits the scope of our conclusions to those concepts present herein.
\section{Methods}\label{sec:methods}

\subsection{Notation}
We describe each word as comprised by form and meaning, which we represent as a pair $(\mathbf{w}^{(n)}, \mathbf{v}^{(n)})$. 
The form $\mathbf{w}^{(n)} \in \Sigma^*$ is represented as a phone string where $\Sigma$ is a phonetic alphabet. 
In this work, we take $\Sigma$ to be the set of 41 phonetic symbols in ASJP plus the end-of-string symbol. 
We write $W$ to denote a $\Sigma^*$-valued random variable. 
The meaning $\mathbf{v}^{(n)} \in \{0, 1\}^{K}$
is represented by a one-hot vector, where $K$ is the number of analysed concepts.\footnote{We note \citet{pimentel2019meaning} used high-dimensional distributional semantic vectors to represent meaning, while we use a one-hot vector. 
However, their work relied on a specific language's \textsc{word2vec}---a choice which could potentially bias our results with that language's properties.
We did, however, run an extra experiment with English \textsc{word2vec}; this led to similar conclusions to the ones presented here.
}
We write $V$ to denote a $\{0, 1\}^K$-valued random variable.
\subsection{Non-Arbitrariness as Mutual Information}

The goal of this work is to measure cross-linguistic form--meaning associations, operationalised
as the \defn{mutual information} (MI) between a form-valued random variable $W$ and a meaning-valued random variable $V$.
Symbolically, we are interested in computing \citep{cover-thomas}:
\begin{equation}\label{eq:mi}
    \MI(W; V) = \Entr(W) - \Entr(W \mid V)
\end{equation}
Intuitively, this quantity captures the uncertainty we have over the form, the entropy $\Entr(W)$, minus how much uncertainty we have over the form \emph{given} the meaning, the conditional entropy $\Entr(W \mid V)$. Thus, if \cref{eq:mi} is zero, its minimum, we have the result that meaning tells us absolutely nothing about the wordform. On the other hand, if \cref{eq:mi} is $\min\{\Entr(W), \Entr(V)\}$,
its maximum, we have that the form is a deterministic function of the meaning (or the opposite; the meaning being deterministically determined given the form).

That the mutual information may take values in $[0, \min\{\Entr(W), \Entr(V)\}]$%
---together with the fact that, for our specific study, $\Entr(W)$ is smaller than $\Entr(V)$---
suggests a more interpretable metric called the \defn{uncertainty coefficient}:
\begin{equation}
    \Unc(W \mid V) = \frac{\MI(W; V)}{\Entr(W)}
\end{equation}
This quantity is the proportion of uncertainty in the form reduced by knowing the meaning. Both mutual information and uncertainty coefficients are general measures of non-arbitrariness. 
One might also inquire about how non-arbitrary a single form--meaning pair is. 
To measure this, we propose \defn{pointwise mutual information} (PMI):
\begin{equation}
    \PMI(\vw; \vv) = \log \frac{p(\vw \mid \vv)}{p(\vw)}
\end{equation}

\subsection{Approximating Mutual Information} \label{sec:approx}

As noted above, we want to estimate the entropy of language agnostic wordforms, i.e.
\begin{equation}
    \Entr(W) = \sum_{\vw \in \wordformspace} p(\vw)  \log \frac{1}{p(\vw)} \label{eq:entropy}
\end{equation}
Unfortunately, we do not know the exact distribution of $p(\mathbf{w})$ and, even if we did, we would need to sum over the infinite set of possible strings $\wordformspace$ to compute this entropy,
which is intractable. 
If we have another probability distribution $\ptheta(\mathbf{w})$, though, we can calculate the cross-entropy between them as an approximation,
i.e.%
\begin{align} \label{eq:cross-entropy}
    \Entr(W) \leq \Entrtheta(W)
     \approx \frac{1}{N} \sum_{n=1}^{N} \log \frac{1}{\ptheta(\tilde{\vw}^{(n)})}
\end{align}
where $\{\tilde{\mathbf{w}}^{(n)}\}_{n=1}^N$ are samples from the true distribution $p$.
Throughout the paper, the tilde marks held-out data, i.e., data not used during model training. 
We note that the approximation becomes exact as $N \rightarrow \infty$ by the weak law of large numbers.  This cross-entropy estimate gives us an upper bound on the actual entropy. This bound is tighter the closer the distributions $p(\mathbf{w})$ and $\ptheta(\mathbf{w})$ are.

\subsection{Estimating the Approximator $\ptheta$} \label{sec:training_q}
How should we train a model to estimate this universal phonotactic distribution $\ptheta(\mathbf{w})$, though?
We train a phone-level language model to predict the next phone given previous ones in a word, i.e.%
\begin{equation}
    \ptheta(\vw) = \prod_{t=1}^{|\vw|} \ptheta(w_t \mid \vw_{<t})
\end{equation}
In this work, we use an LSTM 
as our language model \citep{hochreiter1997long}.
Each phone $w_t$ is represented using a lookup embedding $\mathbf{z}_t \in \mathbb{R}^d$.
These are fed into the LSTM, outputting temporal representations of the sequence:\looseness=-1
\begin{equation}
    \mathbf{h}_t = \LSTM(\mathbf{z}_{t-1}, \mathbf{h}_{t-1})
\end{equation}
where $\mathbf{h}_0$ is the zero vector. These representations are linearly transformed and used in a softmax to approximate the probability distribution:
\begin{equation}
  \ptheta\left(w_{t} \mid  \vw_{< t}\right) = \softmax \left(\mathbf{W}\,\mathbf{h}_t + \mathbf{b}\right)
\end{equation}
All parameters are learned via gradient descent, minimising the cross-entropy in the training set.
\subsection{Cross-linguistic Controls} \label{sec:crosslingual-contros}

As mentioned before, salient regularities between form and meaning across languages might result from large groups of genealogically or spatially related languages. In particular it is practical to consider two independent problems in this respect:

\begin{enumerate}[label=(\roman*)]
    \item \Cref{eq:cross-entropy}'s inequality only holds if $\Entrtheta(W)$ is estimated on a set of datapoints sampled independently from the set of points on which the model $\ptheta$ was trained. As such, the test set should only include languages that are not genealogically or areally related to those in the training set;
    \item Within our dataset, the different size of areal and genealogical groups should be accounted for so that our results are not biased towards particularly large areas or language families.
\end{enumerate}

\paragraph{Train--test split.}
To mitigate the problem referred to in the first item, we cross-validate our models by appealing to the notion of macroareas, large-scale regions of the world that simultaneously maximise internal historical dependency while minimising external ones.
Striking a balance between historical independence and data availability, we consider the following four macroareas: the Americas, Eurasia, Africa, and the Pacific (which in this instantiation includes Papua New Guinea and Australia---see \cref{fig:macroareas}). We will use these macroareas as our folds. Two macroareas will be used at each time for training, while one other is used for validation and the last for testing. Some language families, though, might be present in more than one macroarea (e.g. many European languages are spoken natively in the Americas and Africa). These families will be assigned to the one macroarea which contains most of its family members, since we believe reducing genealogical impact should be preferred over areal impact for our data and purposes, in cases for which such a choice is required.\footnote{As mentioned in \cref{sec:data}, the list of concepts in ASJP was chosen to minimise borrowings across languages. We further note here that loan words are annotated in this dataset and we drop those words for the purpose of our analysis.}
\paragraph{Family size bias.}
The second problem is tackled by weighting each example's contribution to our loss function by the inverse of its family size $l^{(n)}$:
\begin{equation}
  \mathcal{L}\left(\boldsymbol{\theta}\right)
  =\frac{1}{L} \sum_{n=1}^N \frac{1}{l^{(n)}} \log \frac{1}{\ptheta(\vw^{(n)})}
\end{equation}
where $L =\sum_{n=1}^N \frac{1}{l^{(n)}}$ re-normalises the cross-entropy using the family sizes.
This weighted cross-entropy loss function makes per instance contributions of large language families smaller, reducing their impact on the trained model.

To mitigate the same bias effect on the evaluation of validation and test sets, we first get cross-entropies per word. We subsequently average them per language, per family, and per macroarea. This way, each family will have the same effect per macroarea and each macroarea will have the same effect on the overall cross-entropy.

\subsection{Concept-Specific Form Distributions}

We want to compare per-concept phonotactic models with general ones to analyse sound--meaning associations. With that in mind, we condition phone-level language models on meaning:\looseness=-1%
\begin{equation}
    \ptheta(\vw \mid \vv) = \prod\limits_{t=1}^{|\vw|} \ptheta(w_t \mid \vw_{<t}, \vv)
\end{equation}
These models are trained following the same procedures explained above, but conditioning the LSTMs on concept specific representations.
Specifically, the one-hot representation is linearly transformed and fed into the LSTM as its initial state%
\begin{equation}
    \mathbf{h}_0 = \mathbf{W}_0\,\vv
\end{equation}
where the linear transformation $\mathbf{W}_0 \in \mathbb{R}^{d \times K}$
is randomly initialised and learned with the rest of the model.
We then use this distribution to estimate the conditional entropy, analogously to
\cref{eq:cross-entropy}, as in%
\begin{align} \label{eq:cross-entropy_conditioned}
    \Entr(W \mid V) \lesssim \frac{1}{N} \sum_{n=1}^{N} \log \frac{1}{\ptheta(\tilde{\mathbf{w}}^{(n)} \mid \tilde{\mathbf{v}}^{(n)})}
\end{align}
where $\{\tilde{\mathbf{w}}^{(n)}, \tilde{\mathbf{v}}^{(n)}\}_{n=1}^N$ are held-out from--meaning pairs, sampled from the true distribution.%
\footnote{This meaning conditioned model may potentially be better than the raw LSTMs (without conditioning on meaning; due to the extra parameters). To control for this fact, we ran an extra experiment where we estimated $\Entr(W)$ using the meaning dependent model with shuffled concept IDs (so there is no form--meaning association). The results from this shuffled IDs model were very similar to the raw LSTM ones.}

\begin{figure*}[t]
    \centering
    \includegraphics[width=\textwidth]{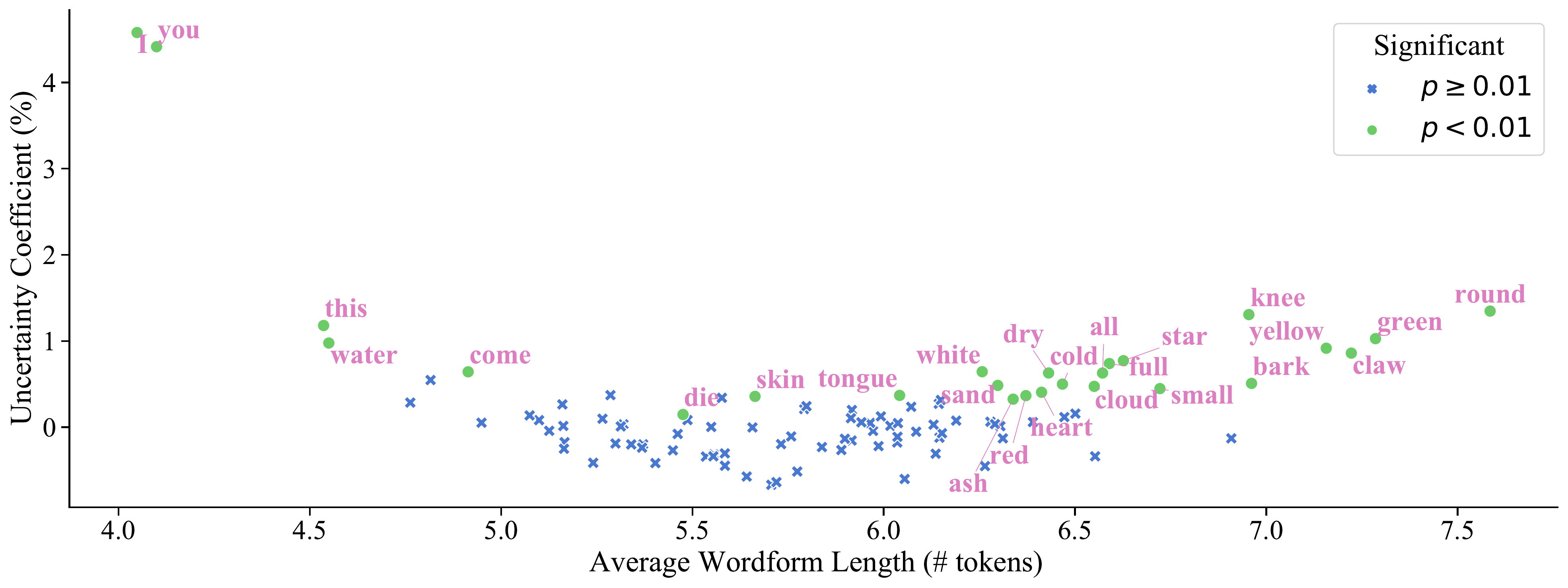}
    \caption{Uncertainty Coefficients for all 100 concepts in ASJP distributed according to their average word length. 
    }
    \label{fig:concepts}
\end{figure*}

\subsection{Non-Arbitrariness as Information}

The mutual information between wordforms and meaning can be decomposed into the difference of two entropy measures.
Unfortunately, we have no way of directly measuring these entropy values without their probability distributions ($p(\mathbf{w})$ and $p(\mathbf{w} \mid \mathbf{v})$). 
We use the estimated cross-entropies as an approximation to this mutual information:%
\begin{align}
\MI(W; V) & = \Entr(W) - \Entr(W \mid V) \\
& \approx \Entrtheta(W) - \Entrtheta(W \mid  V) \label{eq:approx}
\end{align}
We note that \cref{eq:approx} is approximate because it is the difference of two upper bounds.
Furthermore, while there are many ways to estimate mutual information, computing it as the difference between two cross-entropies seems to produce consistent results \citep{pmlr-v108-mcallester20a}.\looseness=-1 

\subsection{Bounds and Optimisation} \label{sec:bayesian}
As mentioned in \cref{sec:approx}, our entropy upper bounds
will be tighter if our models $\ptheta$ better capture $p$.
With this in mind, we optimise the hyper-parameters of our %
models
using Bayesian optimisation with a Gaussian process prior \citep{snoek2012practical}---hyper-parameter ranges are presented in \cref{sec:hyperparameters}.
We train 25 models for each configuration and choose the best one according to the validation set, optimising our weighted cross-entropy loss using AdamW \citep{loshchilov2019decoupled}.

\section{Experiments and Analysis\footnote{Our code is available  at \url{https://github.com/rycolab/form-meaning-associations}.}}

\subsection{Analysis \#1: Overall Mutual Information}\label{sec:overall-results}

We are interested in estimating the cross-linguistic mutual information between meaning and wordforms. With this in mind, we follow the steps described in \cref{sec:training_q}, but instead of only 1 model, we train 25 models using different seeds for each fold (totalling 100 models). Average results---overall and per macroarea---are shown in \cref{tab:celex-full}.
Across macroareas, results indicate a small average contribution of meaning into form (in all cases smaller than $1\%$).\footnote{For comparison, \citet{pimentel2019meaning} estimate intra-language systematicity only accounts for roughly $3 \sim 5\%$ of the entropy in wordforms in English, German and Dutch (given a characteristic sample of the vocabulary).}
A simple permutation test (explained later in this section)
indicates that, under standard levels of significance ($\alpha=0.01$) and after controlling for multiple comparisons,\footnote{All our experiments rely on \newcite{benjamini1995controlling} corrections} this average quantity is significant in 2 out of 4 of the macroareas. 
Nevertheless, this should not be overinterpreted, as unaccounted factors might be responsible for these effects; for instance, the impact of shared history across families in regions smaller than macroareas (almost all human languages have been in contact, directly or indirectly). 
Hence it is reasonable to conclude that there is no definitive evidence for an overall average association at this level of description of the data. We consider specific concept form--meaning associations next.%
\footnote{We ran an experiment changing the macroarea combinations in the train-validation-test sets and the results were stable, leading only to minor numerical changes to \cref{tab:celex-full}.}

\begin{table}[t]
    \centering
    \resizebox{\columnwidth}{!}{%
    \begin{tabular}{l l l c c c}
    \toprule
        \multicolumn{3}{c}{Macroarea} & & \multicolumn{2}{c}{Systematicity}  \\
        \cmidrule(lr{.5em}){1-3}
        \cmidrule(lr{.5em}){5-6}
        Train & Validation & Test
        & $\Entr(W)$
        & $\MI(W ; V)$ & $\Unc(W \mid V)$ \\
    \midrule
P, A$_\mathrm{m}$ & Eurasia & Africa & 3.773 & 0.011$^{*}$ & 0.279\% \\
E, A$_\mathrm{f}$ & Pacific & Americas & 3.901 & 0.007\phantom{$^\dagger$} & 0.173\% \\
A$_\mathrm{f}$, P & Americas & Eurasia & 3.999 & 0.015$^\ddagger$ & 0.376\% \\
A$_\mathrm{m}$, E& Africa & Pacific & 3.755 & 0.016$^\ddagger$ & 0.422\% \\
\midrule
\multicolumn{3}{l}{Average} & 3.857 & 0.012$^\ddagger$ & 0.312\% \\
    \bottomrule 
\multicolumn{3}{l}{$^\ddagger$ $p<0.01$ $^*$ $p<0.1$}
    \end{tabular}
    }
    \caption{Mutual information (in bits per phone) between meaning and wordforms. 
    In the train column, P: Pacific, E: Eurasia, A$_\mathrm{f}$: Africa, A$_\mathrm{m}$: Americas.
    }
    \label{tab:celex-full}
    \vspace{-4.5pt}
\end{table}
\paragraph{Paired Permutation Tests.}
For the permutation test, we first get the average MI over the 25 random seed results for a macroarea. We then permute the signs on these 25 results to create $10^5$ new average MIs. 
By comparing the original result with these permutation ones we get the probability that our MI estimate is significantly larger than zero.
A relevant detail is that these tests are performed on estimates---as opposed to real MI. 
The mutual information is always non-negative, but our estimate is not.
If the MI is zero, we expect our estimates to be negative half the time, since both upper bounds should be roughly equivalent $\Entrtheta(W) \approx \Entrtheta(W \mid V)$.

\paragraph{A note on the LSTMs' quality.} 
Our results strongly rely on the quality of 
approximations. 
Our language independent $\Entr(W)$ estimate is $3.85$ bits per phone.
Meanwhile, the per-language phonotactic cross-entropy found by \citet{pimentel2020phonotactic} 
is, on average, roughly 3 bits per phone%
---generally speaking, these results seem consistent.\footnote{These results are not directly comparable, though, since words are encoded with different phonetic alphabets in ASJP and NorthEuraLex \citep{northeuralex}.}
Furthermore, our model's cross-entropy on the training set is 3.73---while it may have overfit slightly, this is not an aberration.

\begin{figure*}[t]
    \centering
    \includegraphics[width=\textwidth]{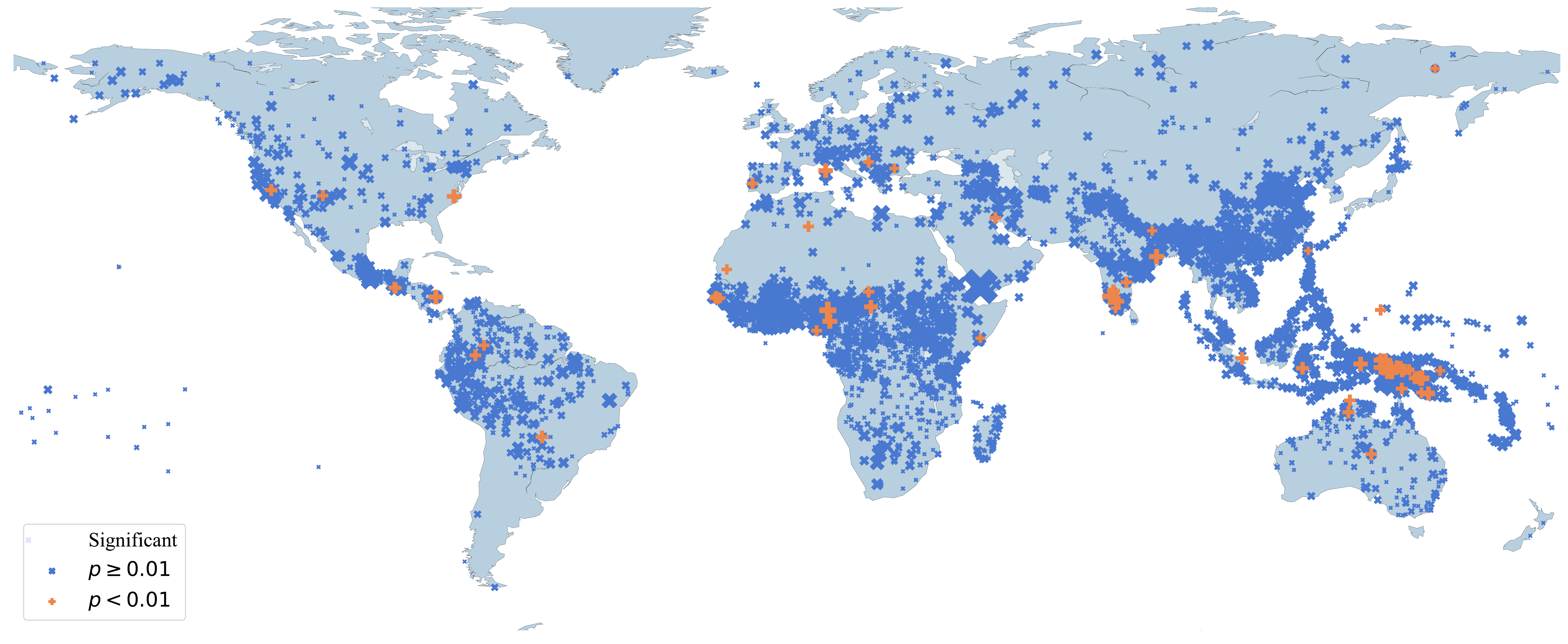}
    \vspace{-22pt}
    \caption{Uncertainty Coefficients for the 5189 languages (9148 doculects) in ASJP. Each language is represented by a point and larger points imply a larger coefficient. Significance was assessed through permutation tests.}
    \label{fig:languages}
\end{figure*}

\subsection{Analysis \#2: Per Concept} \label{sec:concept-analysis}

In this section we focus on concept-specific form--meaning associations. 
With this in mind we group all words for a specific concept $\mathbf{c}\in C$ into a set:
\begin{equation} \label{eq:concept_sets}
    S_\mathbf{c} = \Big\{(\tilde{\vw}^{(n)}, \tilde{\mathbf{v}}^{(n)}) \mid 
    \tilde{\mathbf{v}}^{(n)} = \mathbf{c} \in C\Big\}
\end{equation}
For each such set, we run a permutation test on their approximated pointwise mutual information values $\PMI(\wordform; \mathbf{v}^{(n)})$, assessing if a concept has a statistically significant sound--meaning association.%
\footnote{This permutation test is similar to the one in \cref{sec:overall-results}, but uses the family size corrections discussed in \cref{sec:crosslingual-contros} when averaging results---i.e., for each permutation (and the original one) we average words, languages, families, and macroareas, in this sequence, to get the MI estimate.}
Of the 100 concepts in our dataset, 26 of them have positive mutual information ($p<0.01$). %
This means that, at least in the set of concepts represented in our dataset, non-arbitrary form--meaning associations are not exceptions.

We present the average uncertainty coefficient per concept compared to average wordform length in \cref{fig:concepts}. We do not find any correlation between these measurements.
Analysing these results more closely, we see the pronouns \emph{I} and \emph{you} present the highest coefficient values. Most colours in our dataset (\emph{white}, \emph{red}, \emph{green}, \emph{yellow}) show statistically positive MI. Furthermore, some concepts related to body parts (\emph{tongue}, \emph{skin}, \emph{knee}, \emph{heart}, \emph{claw})  and several concepts related to the environment (\emph{water}, \emph{sand}, \emph{star}, \emph{cloud}, \emph{dry}, \emph{cold}) have statistically positive results.

\citet{wichmann2010sound} also looked at how concepts differ in their degree of form--meaning associations, presenting them in an ordered list together with a measure of how much they deviate from a global average phone usage.
They only look at isolated phone's frequencies, though, and do not control for word length---our mutual information metric controls for both factors. 
When we compare our results to \citepossessive{wichmann2010sound} top 10 list of concepts, we see both contain several body parts (\emph{tongue}, \emph{skin}, \emph{knee}) and pronouns (\emph{I}, \emph{you}).

\subsection{Analysis \#3: Per Language}\label{sec:lang-results}

In their position paper, \newcite{perniss2010iconicity} argue that non-arbitrariness is a general property of language, although sometimes believed to be an exception. They further state that:
\vspace{-5pt}
\begin{displayquote}
``if we look at the lexicon of English (or that of other Indo-European languages), we might be forgiven for thinking that there could be anything but a conventionally determined, arbitrary connection between a given word and its referent. For the vast majority of English words there is an arbitrary relationship between form and meaning.''
\end{displayquote}
\vspace{-5pt}
In fact, in our results we do not find positive MI values, on average, for English. 
In this section, we analyse results per language, trying to find signs of cross-linguistic non-arbitrary associations in them.

Analogously to what we did with concepts, we run permutations tests using the PMIs for the set of words in each language (i.e. sets $S_l$ analogous to $S_{\mathbf{c}}$ in \cref{eq:concept_sets}). 
\Cref{fig:languages} presents the per-language uncertainty coefficient values in a world map.
There are 5189 languages in ASJP, out of those we find that only 85 have significantly positive %
mutual information ($p<0.01$). 
Each language, though, has at most 100 values (the number of concepts), making this a hard statistical test after correcting for the multiple tests.
If we relax our hypothesis testing thresholds to $p<0.05$ (an admittedly much weaker test), then 242 languages present statistically  positive MI---this suggests that, although maybe not common, form--meaning patterns are not a rare exception restricted to a small number of languages.

\begin{table*}[t]
    \centering
    \resizebox{\textwidth}{!}{%
    \begin{tabular}{l c|l c|l c|l c|l c}
    \toprule
        Concept & Tokens & Concept & Tokens & Concept & Tokens & Concept & Tokens & Concept & Tokens \\
    \midrule
blood & s & eye & i & liver & k l r t & path & d t & tree & $\stringending$ \\
bone & s u & fire & $\stringending$ t & louse & m n & say & $\stringending$ & two & r \\
breast & m u & fish & a s & mountain & b d g l o r & see & $\stringending$ e & water & $\stringending$ \\
come & $\stringending$ e & full & l o p t & name & $\stringending$ i & skin & k l p r t & we & $\stringending$ e i n \\
die & t & give & $\stringending$ & neck & o & star & k l o r s t u w & who & $\stringending$ \\
dog & k & horn & k r & new & a & stone & k t & you & $\stringending$ a i n \\
drink & $\stringending$ u & I & $\stringending$ a n & night & N d i l m p r t u & sun & e \\
ear & e l t & knee & N b g k m o r t u & nose & N i u & tongue & d e l n r \\
eat & $\stringending$ & leaf & a l p t & one & k t & tooth & e i \\
    \bottomrule
    \end{tabular}
    }
    \caption{Concept--Token pairs with statistically significant ($p<0.01$ after Benjamini--Hochberg corrections) mutual information in all 4 macro areas. $\stringending$ is the end-of-string token.}
    \label{tab:concept-token-pairs}
\end{table*}

\subsection{Analysis \#4: Per Concept--Token Pair}

We now turn to the relationship between concepts and the phones which appear in them, trying to assess specific concept--phone pairs which present positive MI.
Such a positive value would indicate that concept informs on the presence of that specific phone, suggesting a non-arbitrary association between them.
Similarly to before, we create sets of concept--token pairs:
\begin{align}
    S_{\mathbf{c},s} = & \Big\{(\tilde{w}_t^{(n)}, \tilde{\mathbf{v}}^{(n)}) \mid  \\ 
    & \qquad \quad \tilde{\mathbf{v}}^{(n)}= \mathbf{c} \in C, \tilde{w}_t^{(n)} = s \in \Sigma \Big\}
    \nonumber
\end{align}
where $(\mathbf{c}, s)$ is the analysed concept--token pair and $\tilde{w}_t^{(n)}$ is the $t^\text{th}$ token of word $\tilde{w}^{(n)}$. During this analysis, though, we focus on concept--phone pairs which had statistically significant PMIs in \emph{all four} macroareas, following the controls introduced in \citet{blasi2016sound}, as a way of maximising the chances of finding true history-independent associations (under the risk of increasing the rate of false negatives). With that in mind, we split sets $S_{c,t}$ per macroarea and got the PMI values for each of them, similarly to \cref{sec:concept-analysis}. We threw away pairs which did not occur at least 1000 times together and ran a permutation test with $10^5$ permutations for each concept--token--macroarea tuple. We note a concept--token association does not make a pair probable; the token is simply more likely to appear with the concept than would be without it.

\Cref{tab:concept-token-pairs} presents pairs which were significant in all macroareas ($p<0.01$ after corrections).
After analysis, we find a few interesting results. As mentioned in \cref{sec:introduction}, we see an association between \phon{l} and the concept \concept{tongue} and between \phon{p} and \concept{full}, similarly to \citet{blasi2016sound}. We also see an association between pronouns---e.g. \concept{I}, \concept{we}, \concept{you}---and the end-of-string \phon{$\stringending$}.\footnote{The association of a concept with the \phon{$\stringending$} symbol means the model can more easily predict the end-of-word when conditioned on this concept. This means the length of that concept is not distributed as the average, being more predictable.} This was expected; pronouns are very frequent words in most languages, and such words are usually shorter \citep{zipf1949human}.

As previously found by \newcite{blasi2016sound}, the concept \concept{breast} has a significant association with both \phon{m} and \phon{u}. As they point out, these might be due to the mouth configuration of suckling babies or the sounds they produce when feeding \citep{jakobson1960mama,traunmuller1994sound}. We further find several other pairs which are supported by their findings: \concept{horn}--\phon{k,r}; \concept{knee}--\phon{o,u,k}; \concept{leaf}--\phon{l,p}; \concept{we}--\phon{n}. Furthermore, a nice sanity check is that none of the negative concept--pair associations they found are present in our results.

\subsection{Analysis \#5: Macroareas vs Family}

As a final experiment, we analyse the importance of splitting train--test sets according to macroareas (as discussed in \cref{sec:crosslingual-contros}) in order to minimise areal effects---versus simply splitting languages based on their families.
Even though the list of concepts in ASJP was designed to be resistant to borrowings (and we further remove loan words from our analysis), 
language contacts beyond loan words could still impact results.
One such example is the (potential) impact of Basque in Spanish phonology, which lost word initial /\textipa{f}/ in many words, e.g. \word{hablar}, during the late Middle Ages  \citep[see pg. 91 of][for a longer discussion]{penny1991history}.\looseness=-1

We create 4 folds, splitting them based on glottocode language families, and use 4-fold cross-validation to get family-split results---in opposition to the macroarea-split results. Using family-splits we get an $\MI(W;V)=0.020$ bits, with an uncertainty coefficient of 0.53\% (averaged over the 4-folds)---this is almost twice the overall MI found on the macroarea-splits.
A Welch's $t$-test between both runs shows family-splits have a larger MI than the macroarea results ($p<0.01$), suggesting it is important to control for areal effects when evaluating sound--meaning associations.

\vspace{-1pt}
\section{Conclusion}

In this paper we have provided a holistic assessment of form--meaning associations involving words found in the basic vocabulary in a large number of languages. In agreement with previous findings, we find that on average the meaning does not contribute substantially to the form of the words, but instead the most consistent associations were restricted to a specific subset of all of the words analysed. We find a list of 26 concepts (out of the 100 analysed) with statistically significant form--meaning associations---suggesting that cross-linguistic non-arbitrariness is not a rare exception. Finally, we also find a set of concept--phone pairs with a consistently positive relationship across the four analysed macroareas.

\vspace{-4pt}
\section*{Ethical Considerations} \label{sec:ethical}
\vspace{-2pt}

This paper concerns itself with investigating cross-linguistic form--meaning associations.
We see no direct ethical concerns relating to this work, as it only involves computational experiments on previously collected data.\looseness=-1

\vspace{-4pt}
\section*{Acknowledgments}
\vspace{-2pt}

S{\o}ren Wichmann's research was partly funded by a subsidy from the Russian government to support the Programme of Competitive Development of Kazan Federal University, Russia.
Dami\'{a}n E. Blasi acknowledges funding from the Branco Weiss Fellowship, administered by the ETH Z\"{u}rich.
Dami\'{a}n E. Blasi's research was also executed within the framework of the HSE University Basic Research Program and funded by the Russian Academic Excellence Project `5-100'.

\appendix
\vspace{-4pt}
\section*{Appendix}
\section{Hyper-parameter Optimisation Range} \label{sec:hyperparameters}
\vspace{-2pt}

As mentioned in \cref{sec:bayesian}, we used Bayesian optimisation to tune the model's hyper-parameters.
We consider a log-uniform prior over the embedding size (from $4$ to $1024$), and over the size of the hidden state ($32$ to $1024$). We also considered a uniform prior over the number of layers ($1$ to $4$) and dropout ($0$ to $0.5$).

\bibliography{acl2019}

\begin{thebibliography}{35}
\expandafter\ifx\csname natexlab\endcsname\relax\def\natexlab#1{#1}\fi

\bibitem[{Bankieris and Simner(2015)}]{bankieris2015link}
Kaitlyn Bankieris and Julia Simner. 2015.
\newblock \href
  {https://www.sciencedirect.com/science/article/pii/S0010027714002339} {What
  is the link between synaesthesia and sound symbolism?}
\newblock \emph{Cognition}, 136:186--195.

\bibitem[{Benjamini and Hochberg(1995)}]{benjamini1995controlling}
Yoav Benjamini and Yosef Hochberg. 1995.
\newblock \href {http://www.jstor.org/stable/2346101} {Controlling the false
  discovery rate: A practical and powerful approach to multiple testing}.
\newblock \emph{Journal of the Royal Statistical Society. Series B
  (Methodological)}, 57(1):289--300.

\bibitem[{Bergen(2004)}]{bergen2004psychological}
Benjamin~K. Bergen. 2004.
\newblock \href {https://muse.jhu.edu/article/169798/pdf} {The psychological
  reality of phonaesthemes}.
\newblock \emph{Language}, 80(2):290--311.

\bibitem[{Blasi et~al.(2016)Blasi, Wichmann, Hammarstr{\"o}m, Stadler, and
  Christiansen}]{blasi2016sound}
Dami{\'a}n~E. Blasi, S{\o}ren Wichmann, Harald Hammarstr{\"o}m, Peter~F.
  Stadler, and Morten~H. Christiansen. 2016.
\newblock \href {https://www.pnas.org/content/113/39/10818} {Sound--meaning
  association biases evidenced across thousands of languages}.
\newblock \emph{Proceedings of the National Academy of Sciences of the U.S.A.},
  113(39):10818--10823.

\bibitem[{Cover and Thomas(2012)}]{cover-thomas}
Thomas~M. Cover and Joy~A. Thomas. 2012.
\newblock \emph{Elements of Information Theory}.
\newblock John Wiley \& Sons, New York.

\bibitem[{Dautriche et~al.(2017)Dautriche, Mahowald, Gibson, and
  Piantadosi}]{dautriche2017wordform}
Isabelle Dautriche, Kyle Mahowald, Edward Gibson, and Steven~T. Piantadosi.
  2017.
\newblock \href {https://doi.org/10.1111/cogs.12453} {Wordform similarity
  increases with semantic similarity: An analysis of 100 languages}.
\newblock \emph{Cognitive Science}, 41(8):2149--2169.

\bibitem[{Dellert et~al.(2020)Dellert, Daneyko, M{\"u}nch, Ladygina, Buch,
  Clarius, Grigorjew, Balabel, Boga, Baysarova, M{\"u}hlenbernd, Wahle, and
  J{\"a}ger}]{northeuralex}
Johannes Dellert, Thora Daneyko, Alla M{\"u}nch, Alina Ladygina, Armin Buch,
  Natalie Clarius, Ilja Grigorjew, Mohamed Balabel, Hizniye~Isabella Boga,
  Zalina Baysarova, Roland M{\"u}hlenbernd, Johannes Wahle, and Gerhard
  J{\"a}ger. 2020.
\newblock \href {https://link.springer.com/article/10.1007/s10579-019-09480-6}
  {North{E}ura{L}ex: {A} wide-coverage lexical database of {N}orthern
  {E}urasia}.
\newblock \emph{Language Resources and Evaluation}, 54:273--301.

\bibitem[{Dingemanse(2012)}]{dingemanse2012advances}
Mark Dingemanse. 2012.
\newblock \href {https://onlinelibrary.wiley.com/doi/abs/10.1002/lnc3.361}
  {Advances in the cross‐linguistic study of ideophones}.
\newblock \emph{Language and Linguistics Compass}, 6(10):654--672.

\bibitem[{Dingemanse(2018)}]{dingemanse2018redrawing}
Mark Dingemanse. 2018.
\newblock \href {https://www.glossa-journal.org/articles/10.5334/gjgl.444/}
  {Redrawing the margins of language: {L}essons from research on ideophones}.
\newblock \emph{Glossa: a journal of general linguistics}, 3(1).

\bibitem[{Dingemanse et~al.(2015)Dingemanse, Blasi, Lupyan, Christiansen, and
  Monaghan}]{dingemanse2015arbitrariness}
Mark Dingemanse, Dami{\'a}n~E. Blasi, Gary Lupyan, Morten~H. Christiansen, and
  Padraic Monaghan. 2015.
\newblock \href
  {https://www.sciencedirect.com/science/article/pii/S1364661315001771}
  {Arbitrariness, iconicity, and systematicity in language}.
\newblock \emph{Trends in Cognitive Sciences}, 19(10):603--615.

\bibitem[{Dryer and Haspelmath(2013)}]{wals}
Matthew~S. Dryer and Martin Haspelmath, editors. 2013.
\newblock \href {https://wals.info/} {\emph{WALS Online}}.
\newblock Max Planck Institute for Evolutionary Anthropology, Leipzig.

\bibitem[{Gutierrez et~al.(2016)Gutierrez, Levy, and
  Bergen}]{gutierrez2016finding}
E.~Dario Gutierrez, Roger Levy, and Benjamin Bergen. 2016.
\newblock \href {https://doi.org/10.18653/v1/P16-1225} {Finding non-arbitrary
  form-meaning systematicity using string-metric learning for kernel
  regression}.
\newblock In \emph{Proceedings of the 54th Annual Meeting of the Association
  for Computational Linguistics (Volume 1: Long Papers)}, pages 2379--2388.
  Association for Computational Linguistics.

\bibitem[{Haiman(1980)}]{haiman1980iconicity}
John Haiman. 1980.
\newblock \href {https://www.jstor.org/stable/414448} {The iconicity of
  grammar: {I}somorphism and motivation}.
\newblock \emph{Language}, 56(3):515--540.

\bibitem[{Haynie et~al.(2014)Haynie, Bowern, and LaPalombara}]{haynie2014sound}
Hannah Haynie, Claire Bowern, and Hannah LaPalombara. 2014.
\newblock \href
  {https://journals.plos.org/plosone/article?id=10.1371/journal.pone.0092852}
  {Sound symbolism in the languages of {A}ustralia}.
\newblock \emph{PLoS ONE}, 9(4):e92852.

\bibitem[{Hochreiter and Schmidhuber(1997)}]{hochreiter1997long}
Sepp Hochreiter and J{\"u}rgen Schmidhuber. 1997.
\newblock \href
  {https://direct.mit.edu/neco/article/9/8/1735/6109/Long-Short-Term-Memory}
  {Long short-term memory}.
\newblock \emph{Neural Computation}, 9(8):1735--1780.

\bibitem[{Jakobson(1960)}]{jakobson1960mama}
Roman Jakobson. 1960.
\newblock Why `mama' and `papa'?
\newblock In B.~Kaplan and S.~Wapner, editors, \emph{Perspectives in
  Psychological Theory: Essays in Honor of {H}einz {W}erner}, pages 538--549.
  International Universities Press, New York.

\bibitem[{Johansson and Zlatev(2013)}]{johansson2013motivations}
Niklas Johansson and Jordan Zlatev. 2013.
\newblock \href {https://journals.lub.lu.se/pjos/article/view/9668}
  {Motivations for sound symbolism in spatial deixis: {A} typological study of
  101 languages}.
\newblock \emph{Public Journal of Semiotics}, 5(1):3--20.

\bibitem[{Joo(2019)}]{joo2019phonosemantic}
Ian Joo. 2019.
\newblock \href
  {https://www.degruyter.com/document/doi/10.1515/lingty-2019-0030/html}
  {Phonosemantic biases found in {L}eipzig-{J}akarta lists of 66 languages}.
\newblock \emph{Linguistic Typology}, 24(1):1--12.

\bibitem[{Loshchilov and Hutter(2019)}]{loshchilov2019decoupled}
Ilya Loshchilov and Frank Hutter. 2019.
\newblock \href {https://openreview.net/forum?id=Bkg6RiCqY7} {Decoupled weight
  decay regularization}.
\newblock In \emph{International Conference on Learning Representations}.

\bibitem[{McAllester and Stratos(2020)}]{pmlr-v108-mcallester20a}
David McAllester and Karl Stratos. 2020.
\newblock \href {http://proceedings.mlr.press/v108/mcallester20a.html} {Formal
  limitations on the measurement of mutual information}.
\newblock In \emph{Proceedings of the Twenty Third International Conference on
  Artificial Intelligence and Statistics}, volume 108 of \emph{Proceedings of
  Machine Learning Research}, pages 875--884. PMLR.

\bibitem[{Pache et~al.(2016)Pache, Wichmann, and Zhivlov}]{pache2016words}
Matthias Pache, S{\o}ren Wichmann, and Mikhail Zhivlov. 2016.
\newblock \href {https://benjamins.com/catalog/slcs.173.17pac} {Words for `dog'
  as a diagnostic of language contact in the {A}mericas}.
\newblock In Andrea~L. Berez-Kroeker, Diane~M. Hintz, and Carmen Jany, editors,
  \emph{Language Contact and Change in the {A}mericas: Studies in honour of
  {M}arianne {M}ithun}, pages 385--410. John Benjamins, Amsterdam/Philadelphia.

\bibitem[{Parise et~al.(2014)Parise, Knorre, and Ernst}]{parise2014natural}
Cesare~V. Parise, Katharina Knorre, and Marc~O. Ernst. 2014.
\newblock \href {https://www.pnas.org/content/111/16/6104} {Natural auditory
  scene statistics shapes human spatial hearing}.
\newblock \emph{Proceedings of the National Academy of Sciences of the U.S.A.},
  111(16):6104--6108.

\bibitem[{Penny(2002)}]{penny1991history}
Ralph~John Penny. 2002.
\newblock \href {https://doi.org/10.1017/CBO9780511992827} {\emph{A History of
  the {S}panish Language}}, 2nd ed. edition.
\newblock Cambridge University Press, Cambridge.

\bibitem[{Perniss et~al.(2010)Perniss, Thompson, and
  Vigliocco}]{perniss2010iconicity}
Pamela Perniss, Robin Thompson, and Gabriella Vigliocco. 2010.
\newblock \href
  {https://www.frontiersin.org/articles/10.3389/fpsyg.2010.00227/full}
  {Iconicity as a general property of language: {E}vidence from spoken and
  signed languages}.
\newblock \emph{Frontiers in Psychology}, 1:227.

\bibitem[{Pimentel et~al.(2019)Pimentel, McCarthy, Blasi, Roark, and
  Cotterell}]{pimentel2019meaning}
Tiago Pimentel, Arya~D. McCarthy, Damian Blasi, Brian Roark, and Ryan
  Cotterell. 2019.
\newblock \href {https://doi.org/10.18653/v1/P19-1171} {Meaning to form:
  Measuring systematicity as information}.
\newblock In \emph{Proceedings of the 57th Annual Meeting of the Association
  for Computational Linguistics}, pages 1751--1764, Florence, Italy.
  Association for Computational Linguistics.

\bibitem[{Pimentel et~al.(2020)Pimentel, Roark, and
  Cotterell}]{pimentel2020phonotactic}
Tiago Pimentel, Brian Roark, and Ryan Cotterell. 2020.
\newblock \href {https://doi.org/10.1162/tacl_a_00296} {Phonotactic complexity
  and its trade-offs}.
\newblock \emph{Transactions of the Association for Computational Linguistics},
  8:1--18.

\bibitem[{Romaine(1988)}]{romaine1988pidgins}
Suzanne Romaine. 1988.
\newblock \emph{Pidgin and Creole Languages}.
\newblock Longman, London/New York.

\bibitem[{Saussure(1916)}]{saussure1916course}
Ferdinand~de Saussure. 1916.
\newblock \emph{Course in General Linguistics}.
\newblock Columbia University Press, New York.
\newblock English edition of June 2011, based on the 1959 translation by Wade
  Baskin.

\bibitem[{Shillcock et~al.(2001)Shillcock, Kirby, McDonald, and
  Brew}]{shillcock2001filled}
Richard Shillcock, Simon Kirby, Scott McDonald, and Chris Brew. 2001.
\newblock \href
  {https://www.isca-speech.org/archive_open/archive_papers/diss_01/dis1_053}
  {Filled pauses and their status in the mental lexicon}.
\newblock In \emph{ISCA Tutorial and Research Workshop (ITRW) on Disfluency in
  Spontaneous Speech}.

\bibitem[{Snoek et~al.(2012)Snoek, Larochelle, and Adams}]{snoek2012practical}
Jasper Snoek, Hugo Larochelle, and Ryan~Prescott Adams. 2012.
\newblock \href
  {https://proceedings.neurips.cc/paper/2012/file/05311655a15b75fab86956663e1819cd-Paper.pdf}
  {Practical {B}ayesian optimization of machine learning algorithms}.
\newblock In \emph{Advances in Neural Information Processing Systems},
  volume~25. Curran Associates, Inc.

\bibitem[{Swadesh(1955)}]{swadesh1955towards}
Morris Swadesh. 1955.
\newblock \href {https://www.jstor.org/stable/1263939} {Towards greater
  accuracy in lexicostatistic dating}.
\newblock \emph{International Journal of American Linguistics}, 21(2):121--137.

\bibitem[{Traunm{\"u}ller(1994)}]{traunmuller1994sound}
Hartmut Traunm{\"u}ller. 1994.
\newblock Sound symbolism in deictic words.
\newblock In Hans Auli and Peter af~Trampe, editors, \emph{Tongues and Texts
  Unlimited. Studies in Honour of {T}ore {J}ansson on the Occasion of his
  Sixtieth Anniversary}, pages 213--234. Dept. of Classical Languages,
  Stockholm University, Stockholm.

\bibitem[{Wichmann et~al.(2010)Wichmann, Holman, and Brown}]{wichmann2010sound}
S{\o}ren Wichmann, Eric~W. Holman, and Cecil~H. Brown. 2010.
\newblock \href {https://www.mdpi.com/1099-4300/12/4/844} {Sound symbolism in
  basic vocabulary}.
\newblock \emph{Entropy}, 12(4):844--858.

\bibitem[{Wichmann et~al.(2020)Wichmann, Holman, and Brown}]{wichmann2020asjp}
S{\o}ren Wichmann, Eric~W. Holman, and Cecil~H. Brown. 2020.
\newblock \href {https://asjp.clld.org/} {The {ASJP} database (version 19)}.
\newblock Accessed on April 2, 2020.

\bibitem[{Zipf(1949)}]{zipf1949human}
George~Kingsley Zipf. 1949.
\newblock \emph{Human Behavior and the Principle of Least Effort}.
\newblock Addison-Wesley Press, Cambridge, Mass.

\end{thebibliography}
\bibliographystyle{acl_natbib}

\end{document}